\documentclass[11pt]{article}

\usepackage{epsfig}
\usepackage{theorem}
\usepackage{enumerate}
\usepackage{apa-uiuc}
\usepackage{amsmath}
\usepackage{amssymb}
\usepackage{boxedminipage}
\usepackage{subfigure}
\usepackage{url}

\begin{document}
\begin{titlepage}
\oddsidemargin 6mm
\vspace*{2.0in}
\vspace*{2ex}
\begin{center}
{\bf Fitness Inheritance in the \\Bayesian Optimization Algorithm}
 
\addvspace{0.25in}
{\bf Martin Pelikan and Kumara Sastry}\\
\addvspace{0.4in}
IlliGAL Report No. 2004009 \\
February 2004 \\
 
\vspace*{2.8in}
Illinois Genetic Algorithms Laboratory \\
University of Illinois at Urbana-Champaign \\
117 Transportation Building \\
104 S. Mathews Avenue
Urbana, IL 61801 \\
Office:  (217) 333-2346\\
Fax: (217) 244-5705 \\
\end{center}
\end{titlepage}

\title{Fitness Inheritance in the \\Bayesian Optimization Algorithm}

\author{Martin Pelikan\\
Dept. of Math. and Computer Science, 320 CCB\\
University of Missouri at St. Louis\\
8001 Natural Bridge Rd.,
St. Louis, MO 63121\\
\url{pelikan@cs.umsl.edu}\\
~\\
Kumara Sastry\\
Illinois Genetic Algorithms Laboratory, 107 TB\\
University of Illinois at Urbana-Champaign\\
104 S. Mathews Ave.
Urbana, IL 61801\\
\url{kumara@illigal.ge.uiuc.edu}
}


\maketitle 


\begin{abstract}
This paper describes how fitness inheritance can be used to estimate fitness for a proportion of newly sampled candidate solutions in the Bayesian optimization algorithm (BOA). The goal of estimating fitness for some candidate solutions is to reduce the number of fitness evaluations for problems where fitness evaluation is expensive. Bayesian networks used in BOA to model promising solutions and generate the new ones are extended to allow not only for modeling and sampling candidate solutions, but also for estimating their fitness. The results indicate that fitness inheritance is a promising concept in BOA, because population-sizing requirements for building appropriate models of promising solutions lead to good fitness estimates even if only a small proportion of candidate solutions is evaluated using the actual fitness function. This can lead to a reduction of the number of actual fitness evaluations by a factor of 30 or more.
\end{abstract}


\section{Introduction}
To ensure reliable convergence to a global optimum, genetic and evolutionary algorithms (GEAs) must often maintain a large population of candidate solutions for a number of iterations. However, in many real-world problems, fitness evaluation is computationally expensive and evaluating even moderately sized populations of candidate solutions is intractable. For example, fitness evaluation may include a large finite element analysis, it may consist of a complex traffic simulation, or it may require interaction with a human expert. 

This leads to an interesting question: Would it be possible to make GEAs evolve not only the population of candidate solutions, but also a model of fitness, which could be used to evaluate a certain proportion of newly generated candidate solutions (fitness inheritance)? Fortunately, the answer to the above question is positive, and a few studies have been made to support this argument. Methods were proposed for fitness inheritance in the simple genetic algorithm (GA)~\cite{SmithR:94} and the univariate marginal distribution algorithm (UMDA)~\cite{Sastry:01}. In both cases, the results were promising and suggested that fitness inheritance can significantly reduce the number of fitness evaluations. 

The purpose of this paper is to propose a method that uses models of promising solutions developed by the Bayesian optimization algorithm (BOA)~\cite{Pelikan:99a,Pelikan:thesis} to model the fitness landscape and estimate fitness of newly generated candidate solutions. Two types of models are considered: (1) traditional Bayesian networks with full conditional probability tables (CPTs) used in BOA and (2) Bayesian networks with local structures used in BOA with decision graphs (dBOA)~\cite{Pelikan:01a*} and the hierarchical BOA (hBOA)~\cite{Pelikan:01*,Pelikan:03b}. Since the model in BOA captures significant nonlinearities in the fitness landscape, using this model as the basis for developing a model of the fitness landscape seems to be a promising approach. Of course, other methods, such as neural networks or various regression models, could be used instead. The proposed method is examined on BOA with decision trees on three example problems: onemax, concatenated traps of order 4, and concatenated traps of order 5. The results indicate that fitness inheritance is beneficial in BOA even if only less than $1\%$ of candidate solutions are evaluated using the actual fitness function. It turns out that due to the population sizing requirements for creating a correct model of promising solutions, the more fitness inheritance, the better.

The paper starts by discussing BOA and previous fitness inheritance studies. Section~\ref{section-BOA-inheritance} presents the proposed method for fitness inheritance in BOA. Section~\ref{section-experiments} presents and discusses experimental results. Section~\ref{section-conclusions} summarizes and concludes the paper.


\section{Bayesian optimization algorithm}
Probabilistic model-building genetic algorithms (PMBGAs)~\cite{Pelikan:02} replace traditional variation operators of genetic and evolutionary algorithms~\cite{Holland:75a,Goldberg:89d} by building a probabilistic model of promising solutions and sampling the model to generate new candidate solutions. The Bayesian optimization algorithm (BOA)~\cite{Pelikan:99a} uses Bayesian networks to model candidate solutions. 

BOA evolves a population of candidate solutions to the given problem. The first population of candidate solutions is usually generated randomly according to a uniform distribution over all solutions. The population is updated for a number of iterations using two basic operators: (1) selection, and (2) variation. The selection operator selects better solutions at the expense of the worse ones from the current population, yielding a population of promising candidates. The variation operator starts by learning a probabilistic model of the selected solutions that encodes features of these promising solutions and the inherent regularities. Bayesian networks are used to model promising solutions because Bayesian networks are among the most powerful tools for capturing and representing decomposition, which is an inherent feature of most complex real-world systems. The variation operator then proceeds by sampling the probabilistic model to generate new solutions, which are incorporated into the original population. Here, a simple replacement scheme is used where new solutions fully replace the original population. A more detailed description of BOA can be found in~\citeN{Pelikan:thesis}.

The remainder of this section discusses Bayesian networks, which are going to serve as the basis for developing the model of fitness in BOA.


\subsection{Bayesian Networks}

Bayesian networks (BNs)~\cite{Howard:81,Pearl:88,Buntine:91} are among the most popular graphical models, where statistics, modularity, and graph theory are combined in a practical tool for estimating probability distributions and inference. A Bayesian network is defined by two components: (1) a structure, and (2) parameters. The structure is encoded by a directed acyclic graph with the nodes corresponding to the variables in the modeled data set (in this case, to the positions in solution strings) and the edges corresponding to conditional dependencies. The parameters are represented by a set of conditional probability tables (CPTs) specifying a conditional probability for each variable given any instance of the variables that the variable depends on. 

A Bayesian network encodes a joint probability distribution\index{distribution!joint} given by
\begin{equation}
\label{eq-joint-distribution}
p(X) = \prod_{i=1}^{n}{p(X_i| \Pi_i)},
\end{equation}
where $X=(X_0,\ldots,X_{n-1})$ is a vector of all the variables in the problem; $\Pi_i$ is the set of parents of $X_i$ (the set of nodes from which there exists an edge to $X_i$); and $p(X_i| \Pi_i)$ is the conditional probability of $X_i$ given its parents $\Pi_i$.
                                                                                      
A directed edge relates the variables so that in the encoded distribution, the variable corresponding to the terminal  node is conditioned on the variable corresponding to
the initial node. More incoming edges into a node result in a conditional probability
of the variable with a condition containing all its parents. In addition to encoding dependencies, each Bayesian network encodes a set of independence assumptions. Independence assumptions state that each variable is independent of any of its antecedents in
the ancestral ordering, given the values of the variable's parents.

To learn Bayesian networks, a greedy algorithm is usually used for its efficiency and robustness. The greedy algorithm starts with an empty Bayesian network. Each iteration then adds an edge into the network that improves quality of the network the most. Network quality can be measured by any popular scoring metric for Bayesian networks, such as the Bayesian Dirichlet metric with likelihood equivalence (BDe)~\cite{Cooper:92,Heckerman+al:94} or the Bayesian information criterion (BIC)~\cite{Schwarz78a}. The learning is terminated when no more improvement is possible.

\subsection{Conditional probability tables (CPTs)}
Conditional probability tables (CPTs) store conditional probabilities $p(X_i|\Pi_i)$ for each variable $X_i$. The number of conditional probabilities for a variable that is conditioned on $k$ parents grows exponentially with $k$. For binary variables, for instance, the number of conditional probabilities is $2^k$, because there are $2^k$ instances of $k$ parents and it is sufficient to store the probability of the variable being $1$ for each such instance. Figure~\ref{figure-decision-tree-and-graph} shows an example CPT for $p(X_1|X_2,X_3,X_4)$.

Nonetheless, the dependencies sometimes also contain regularities. Furthermore, the exponential growth of full CPTs often obstructs the creation of models that are both accurate and efficient. That is why Bayesian networks are often extended with local structures that allow more efficient representation of local conditional probability distributions than full CPTs~\cite{Chickering:97,Friedman:99}. 

\subsection{Decision trees and graphs for conditional probabilities}

Decision trees are among the most flexible and efficient local structures, where conditional probabilities of each variable are stored in one decision tree. Each internal (non-leaf) node in the decision tree for $p(X_i|\Pi_i)$ has a variable from $\Pi_i$ associated with it and the edges connecting the node to its children stand for different values of the variable. For binary variables, there are two edges coming out of each internal node; one edge corresponds to 0, whereas the other edge corresponds to 1. For more than two values, either one edge can be used for each value, or the values may be classified into several categories and each category would create an edge. 

\begin{figure}[t]
\begin{center}
  \subfigure[Conditional probability table.]{~~~~~~~~~~~~~\epsfig{file=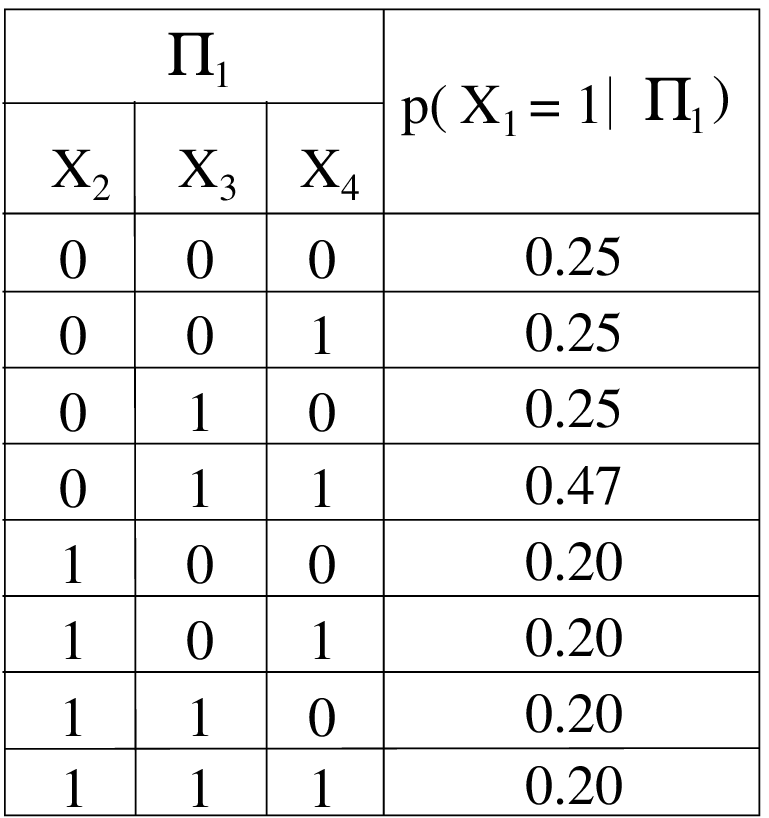,height=1.6in}~~~~~~~~~~}\hspace{0in}
\subfigure[Decision tree.]{\epsfig{file=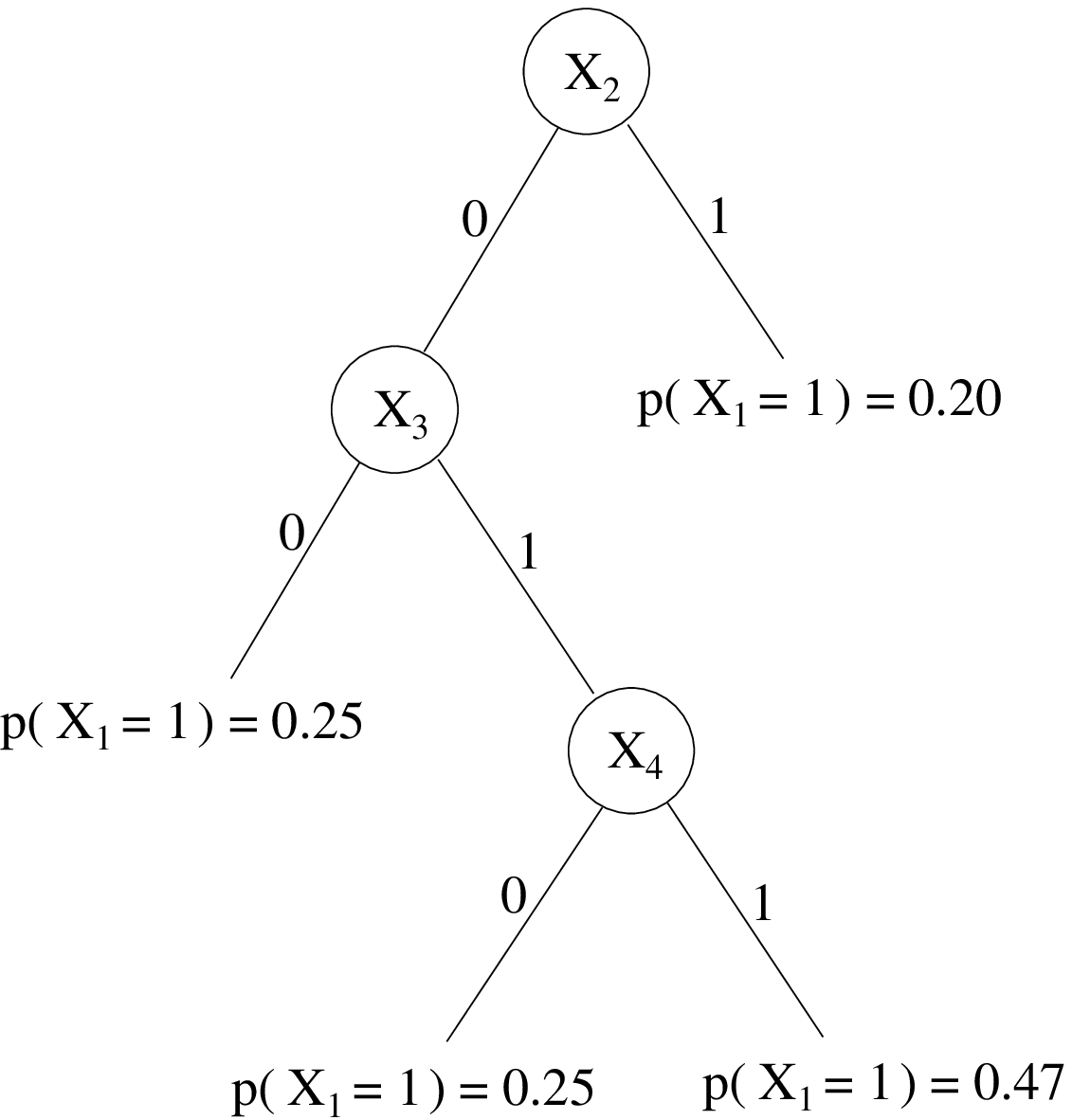,height=1.6in}}\hspace{0in}
\subfigure[Decision graph.]{~~~~~~~\epsfig{file=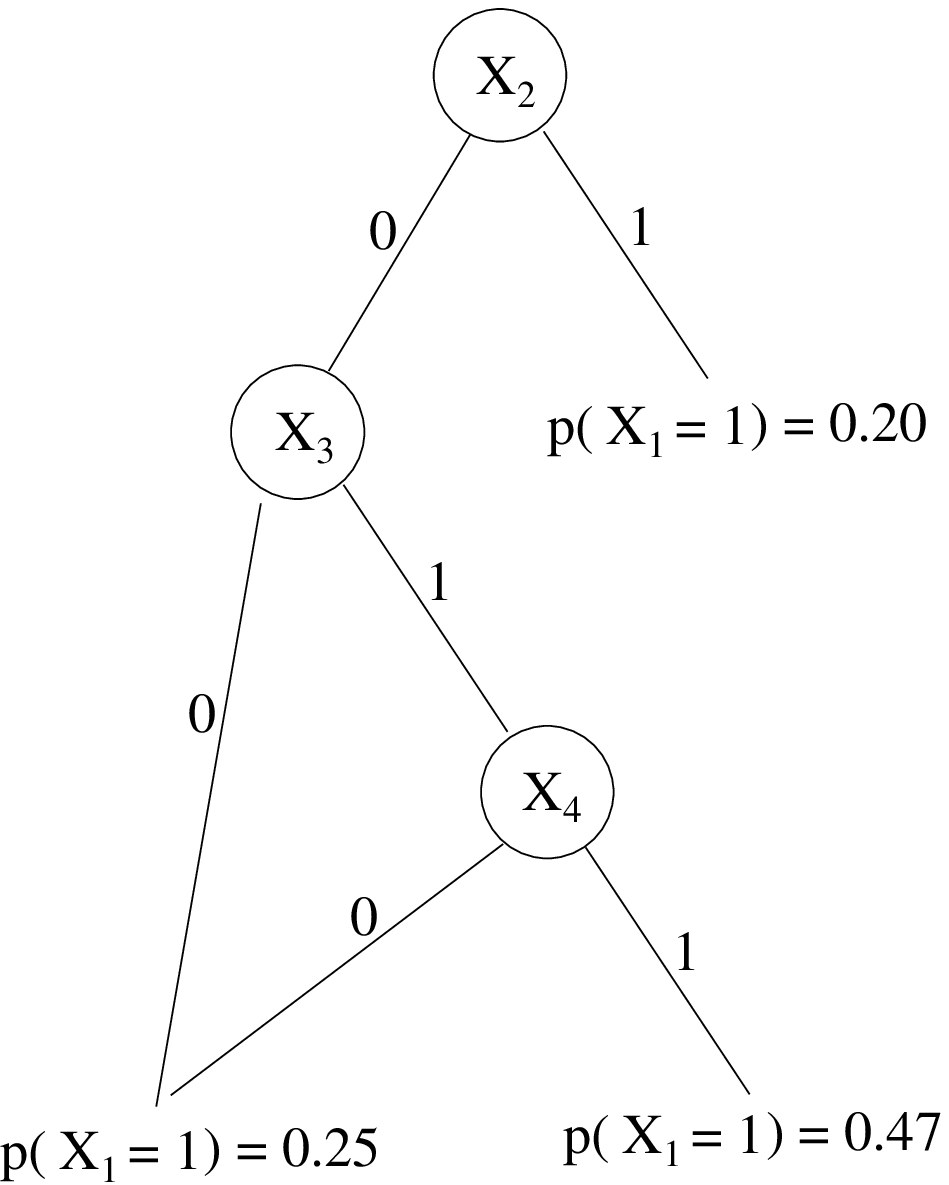,height=1.6in}}
\end{center}
\vspace*{-3ex}
\caption{A conditional probability table for $p(X_1|X_2,X_3,X_4)$ using traditional representation (a) as well as local structures (b and c).}
\vspace*{-2ex}
\label{figure-decision-tree-and-graph}
\end{figure}

Each path in the decision tree for $p(X_i|\Pi_i)$ that starts in the root of the tree and ends in a leaf encodes a set of constraints on the values of variables in $\Pi_i$. Each leaf stores the value of a conditional probability of $X_i=1$ given the condition specified by the path from the root of the tree to the leaf. A decision tree can encode the full conditional probability table for a variable with $k$ parents if it splits to $2^k$ leaves, each corresponding to a unique condition. However, a decision tree enables more efficient and flexible representation of local conditional distributions. See Figure~\ref{figure-decision-tree-and-graph}b for an example decision tree for the conditional probability table presented earlier.

A decision graph allows more edges to terminate in a single node. In other words, internal nodes in the decision tree are allowed to share children and, as a result, each node can have more than one parent. That makes this representation even more flexible. However, our experience indicates that, in BOA, decision graphs usually do not provide better performance than decision trees. See Figure~\ref{figure-decision-tree-and-graph}c for an example decision graph.

To learn Bayesian networks with decision trees, a decision tree for each variable $X_i$ is initialized to an empty tree with a univariate probability of $X_i=1$. In each iteration, each leaf of each decision tree is split to determine how quality of the current network improves by executing the split, and the best split is performed. The learning is finished when no splits improve the current network anymore. Quality of each model can be estimated using any popular scoring metric. Here we use a combination of the BDe~\cite{Cooper:92,Heckerman+al:94} and BIC~\cite{Schwarz78a} metrics, where the BDe score is penalized with the number of bits required to encode parameters~\cite{Pelikan:thesis}. For decision graphs, a merge operation is introduced to allow for merging two leaves of any (but always the same) decision graph.

\section{Previous fitness inheritance studies}



Despite the importance of fitness inheritance in robust population-based search, surprisingly few studies of fitness inheritance can be found. This section reviews the most important studies.

\subsection{Fitness inheritance in the simple GA}

\citeN{SmithR:94} proposed two approaches to fitness inheritance in the simple GA~\cite{Goldberg:89d}. The first approach is to compute the fitness of an offspring as the average fitness of its parents. The second approach is to consider a weighted average based on how similar the offspring is to each parent. The results indicated that GAs with fitness inheritance outperformed those without inheritance. However, the above study of fitness inheritance did not consider the effects of fitness inheritance on crucial GA parameters such as the population size and the number of generations. As a result, the speed-up achieved by using fitness inheritance could not be estimated properly. 

\citeN{Zheng:97} used the aforementioned fitness inheritance model in the simple GA for design of vector quantization codebooks. 

\subsection{Fitness inheritance in PMBGAs}
\citeN{Sastry:01} considered the univariate marginal distribution algorithm (UMDA), which is one of the simplest PMBGAs. Using fitness inheritance in UMDA introduces new challenges, because UMDA does not use two-parent recombination and therefore it is difficult to find direct correspondence between parents and their offspring. Instead, Sastry et al. extend the probabilistic model to allow for estimating fitness of newly sampled candidate solutions. 

UMDA models the population of promising solutions after selection using the probability vector, which stores the probability of a 1 at each position. These probabilities are then used to sample new candidate solutions. To incorporate fitness inheritance, the probability vector $p=(p_1, p_2, \ldots, p_n)$ is extended to include additional two statistics $\bar{f}(X_i=0)$ and $\bar{f}(X_i=1)$ for each string position $i$. The term $\bar{f}(X_i=0)$ denotes the average fitness  of all solutions where the $i$th bit is $0$; analogously, the term $\bar{f}(X_i=1)$ denotes the average fitness of solutions with the $i$th bit equal to $1$. The fitness of each new solution can then estimated as
\begin{equation}
\label{eq-umda-estimate}
f_{est}(X_1, X_2, \ldots, X_n) = \bar{f} + \sum_{i=1}^n \left( \bar{f}(X_i) - \bar{f} \right),
\end{equation}
where $\bar{f}$ is the average fitness of all solutions used to estimate the fitness.

\shortciteN{Sastry:01} also developed theory for fitness inheritance in UMDA on onemax that estimates the number of actual fitness evaluations when a given proportion of candidate solutions inherits fitness, whereas the remaining candidate solutions are evaluated using the actual fitness. The basic idea is to start by adapting the population sizing and time-to-convergence models to UMDA with fitness inheritance, and relate these quantities to their counterparts in standard UMDA. If optimal population size is used in both cases, Sastry et al. showed that only about $20\%$ evaluations can be saved. However, if the same population size is used in both cases, the number of evaluations decreases by a factor of more than three.


\section{Modeling fitness in BOA}
\label{section-BOA-inheritance}

This section describes how the fitness model is built and updated using Bayesian networks, and how new candidate solutions can be evaluated using the model. Both Bayesian networks with full CPTs as well as the ones with local structures are discussed. The section also discusses where the statistics can be acquired from to built an accurate fitness model.

\subsection{Modeling fitness using Bayesian networks}

In UMDA, probabilities of a $1$ at each position that form the  probability vector are each coupled with an average fitness of a 0 and a 1 at that position. Analogically, Bayesian networks can be extended to incorporate an average fitness of a 0 and a 1 for each statistic stored by the model. 

In BOA, for every variable $X_i$ and each possible value $x_i$ of $X_i$, an average fitness of solutions with $X_i=x_i$ must be stored for each instance $\pi_i$ of $X_i$'s parents $\Pi_i$. In the binary case, each row in the conditional probability table is thus extended by two additional entries. Figure~\ref{figure-decision-tree-and-graph-with-inheritance}a shows an example conditional probability table extended with fitness information based on the conditional probability table presented in Figure~\ref{figure-decision-tree-and-graph}a. The fitness can then be estimated as
\begin{equation}
f_{est}(X_1, X_2, \ldots, X_n) = \bar{f} + \sum_{i=1}^n \left( \bar{f}(X_i|\Pi_i) - \bar{f}(\Pi_i) \right),
\end{equation}
where $\bar{f}(X_i|\Pi_i)$ denotes the average fitness of solutions with $X_i$ and $\Pi_i$, and $\bar{f}(\Pi_i)$ is the average fitness of all solutions with $\Pi_i$. Clearly,
\begin{equation}
\bar{f}(\Pi_i) = \sum_{X_i} p(X_i|\Pi_i) \bar{f}(X_i|\Pi_i).
\end{equation}

\subsection{Modeling fitness using Bayesian networks with decision graphs}

\begin{figure}[t]
\begin{center}
  \subfigure[Conditional probability table.]{~~~~\epsfig{file=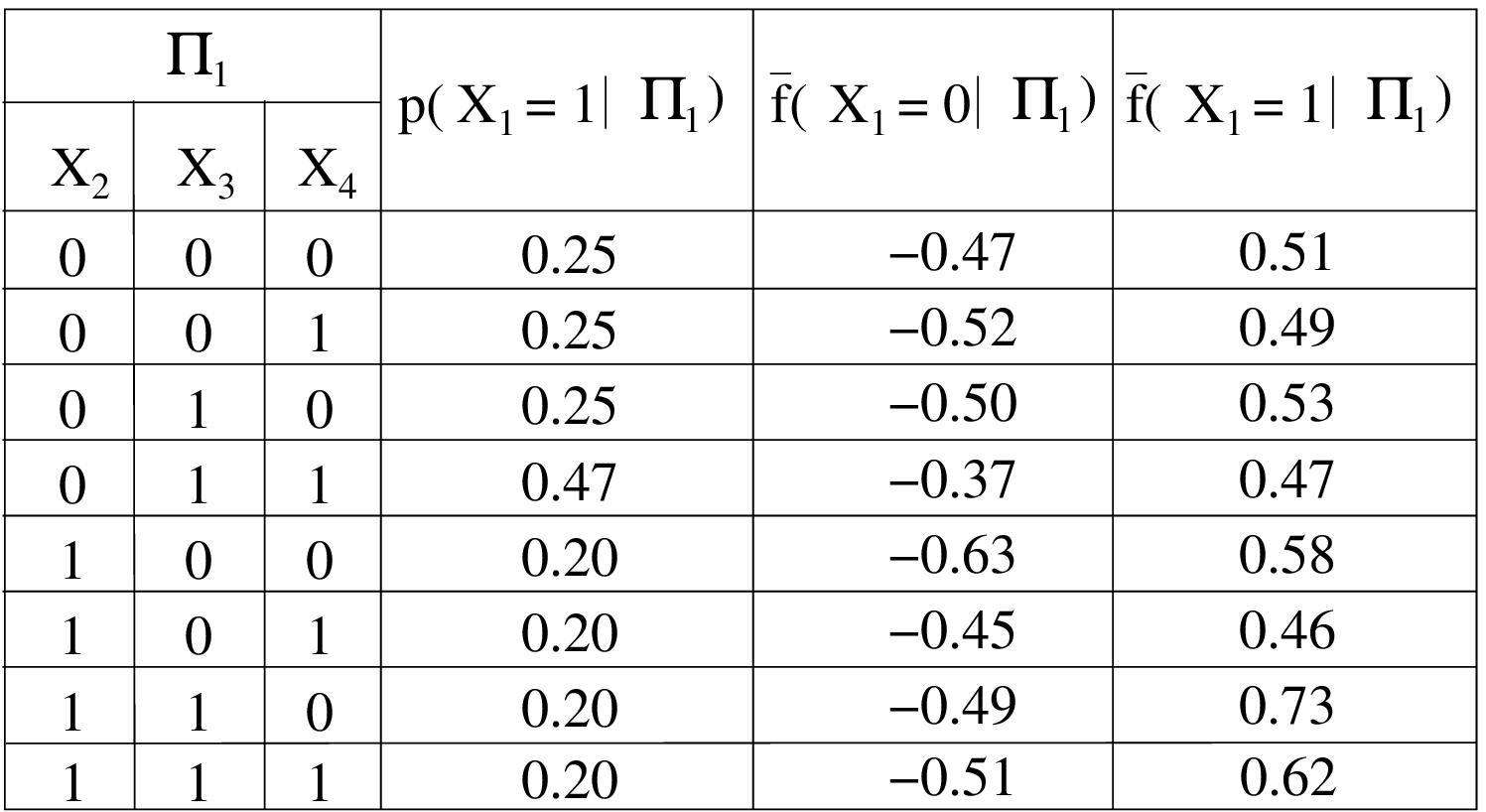,height=1.6in}~~~}
\subfigure[Decision tree.]{~~\epsfig{file=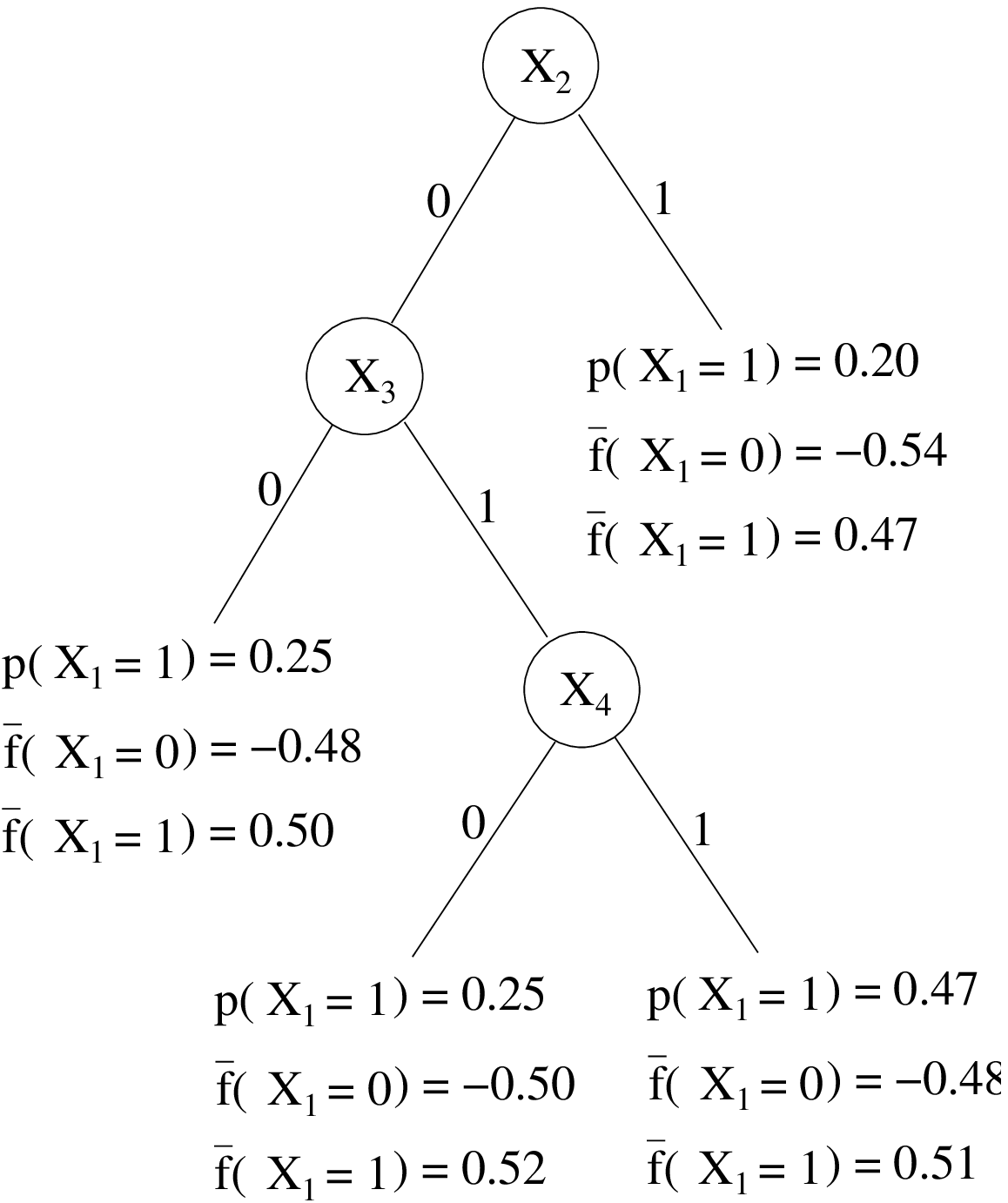,height=1.6in}~}\hspace{0in}
\subfigure[Decision graph.]{~~~~~\epsfig{file=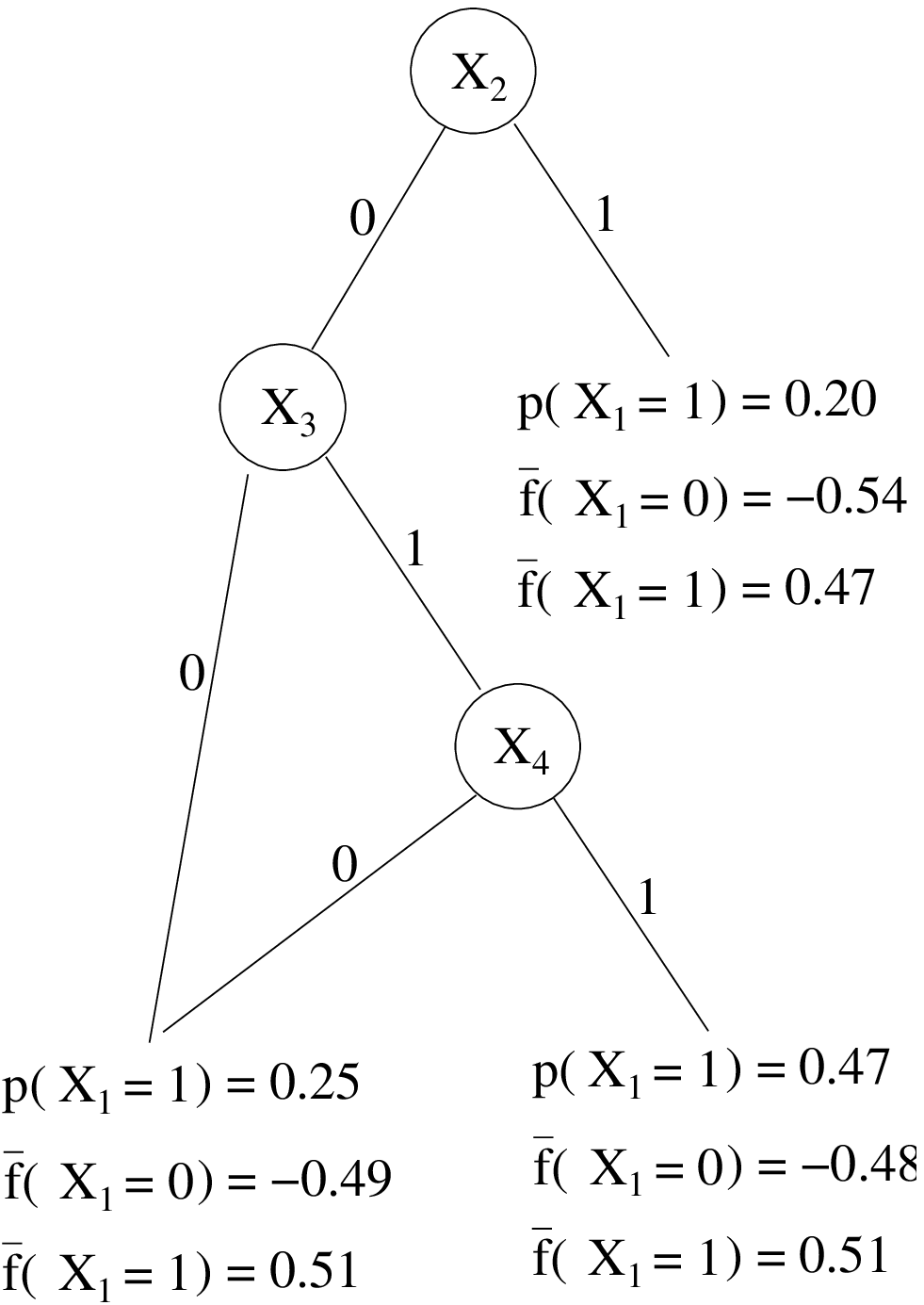,height=1.6in}~~~~}
\end{center}
\vspace*{-3ex}
\caption{Fitness inheritance in a conditional probability table for $p(X_1|X_2,X_3,X_4)$ (a) and its representation using local structures (b and c).}
\label{figure-decision-tree-and-graph-with-inheritance}
\vspace*{-2ex}
\end{figure}

A similar method as for full CPTs can be used to incorporate fitness information into Bayesian networks with decision trees or graphs. The average fitness of each instance of each variable must be stored in every leaf of a decision tree or graph. Figure~\ref{figure-decision-tree-and-graph-with-inheritance} shows an example decision tree and graph extended with fitness information based on the decision tree and graph presented earlier in Figure~\ref{figure-decision-tree-and-graph}. The fitness averages in each leaf are restricted to solutions that satisfy the condition specified by the path from the root of the tree to the leaf. 

\subsection{Where to inherit fitness from?}

We still have not faced the following question: Where to obtain information to compute statistics used for fitness inheritance? More specifically, for each instance $x_i$ of $X_i$ and each instance $\pi_i$ of $X_i$'s parents $\Pi_i$, we must compute the average fitness of all solutions with $X_i=x_i$ and $\Pi_i=\pi_i$. Here we use two sources for computing the fitness-inheritance statistics:

\vspace{-1.95ex}
\begin{enumerate}
\item Selected parents that were evaluated using the actual fitness function, and 
\item the offspring that were evaluated the actual fitness function.
\end{enumerate}

The reason for restricting computation of fitness-inheritance statistics to selected parents and offspring is that the probabilistic model used as the basis for selecting relevant statistics represents nonlinearities in the population of parents and the population of offspring. Since it is best to maximize learning data available, it seems natural to use these two populations to compute the fitness-inheritance statistics. The reason for restricting input for computing these statistics to solutions that were evaluated using the actual fitness function is that the fitness of other solutions was estimated only and it involves errors that could mislead fitness inheritance and propagate through generations. Both using only those solutions that were evaluated using the actual fitness function and incorporating the offspring in estimating inheritance statistics differs from previous fitness inheritance studies~\cite{SmithR:94,Sastry:01}.


\section{Experiments}
\label{section-experiments}

This section describes experiments and provides experimental results. Test problems are described first. Next, experimental results are presented and discussed.

\subsection{Onemax}
Onemax is a simple linear function that computes the sum of bits in the input binary string:
\begin{equation}
f_{onemax}(X_1, X_2, \ldots, X_n) = \sum_{i=1}^n X_i,
\end{equation}
where $(X_1, X_2, \ldots, X_n)$ denotes the input binary string of $n$ bits. In onemax, the fitness contribution of each bit is independent of its context. That is why a simple model used in UMDA that considers each variable independently of other variables suffices and yields convergence to the optimum in about $O(n\log n)$ evaluations. However, any other models of bounded complexity should work well, and practically any crossover operator used in standard GAs should also suffice. 

In the model of fitness developed by BOA, the average fitness of a $1$ in any leaf should be approximately $0.5$, whereas the average fitness of a $0$ in any leaf should be approximately $-0.5$. As a result, solutions will get penalized for $0$s, while they would be rewarded for $1$s. The average fitness will vary throughout the run. This paper considers onemax of $n=50$ bits.

\subsection{Concatenated 4-bit trap}

In concatenated 4-bit traps~\cite{Ackley:87b,Deb:94b}, the input string is first partitioned into independent groups of $4$ bits each. This partitioning should be unknown to the algorithm, but it should not change during the run. A 4-bit trap function is applied to each group of 4 bits and the contributions of all traps are added together to form the fitness. Each 4-bit trap is defined as follows:
\begin{equation}
trap_4(u) = 
\left\{
\begin{array}{ll}
4 & \mbox{~~if $u=4$} \\
3-u & \mbox{~~otherwise}
\end{array}
\right.,
\end{equation}
where $u$ is the number of $1$s in the input string of $4$ bits. An important feature of traps is that in each of the 4-bit traps, all 4 bits must be treated together, because all statistics of lower order lead the algorithm away from the optimum. That is why most crossover operators as well as the model in UMDA will fail at solving this problem faster than in exponential number of evaluations, which is just as bad as blind search. 

Unlike in onemax, $\bar{f}(X_i=0)$ and $\bar{f}(X_i=1)$ depend on the state of the search because the distribution of contexts of each bit changes over time and bits in a trap are not independent. The context of each leaf also determines whether  $\bar{f}(X_i=0)<\bar{f}(X_i=1)$ or $\bar{f}(X_i=0)>\bar{f}(X_i=1)$ in the leaf. This paper considers a trap consisting of 10 copies of the 4-bit trap, where the total number of bits is $n=40$. 

\subsection{Concatenated 5-bit trap}
Concatenated traps of order 5 can be defined analogically to traps of order 4, but instead of dealing with groups of 4 bits, groups of 5 bits are considered. The contribution of each group of $5$ bits is computed as
\begin{equation}
trap_5(u) = 
\left\{
\begin{array}{ll}
5 & \mbox{~~if $u=5$} \\
4-u & \mbox{~~otherwise}
\end{array}
\right.,
\end{equation}
where $u$ is the number of $1$s in the input string of $5$ bits. Traps of order 5 also necessitate that all bits in each group are treated together, because statistics of lower order are misleading. 

Average fitness values $\bar{f}(X_i)$ depend on context similarly as for traps of order 4, and they thus follow similar dynamics. This paper considers a trap consisting of 10 copies of the 5-bit trap, where the total number of bits is $n=50$. 

\subsection{Experimental results}

On each test problem, the following fitness inheritance proportions were considered: $0$ to $0.9$ with step $0.1$, $0.91$ to $0.99$ with step $0.01$, and $0.991$ to $0.999$ with step $0.001$. For each test problem and fitness inheritance proportion, 30 independent experiments were performed. Each experiment consisted of 10 independent runs with the minimum population size to ensure convergence to a solution within $10\%$ of the optimum (i.e., with at least $90\%$ correct bits) in all 10 runs. For each experiment, bisection method was used to determine the minimum population size, and the number of evaluations (excl. the evaluations done using the model of fitness) was recorded. The average of 10 runs in all experiments was then computed and displayed as a function of the proportion of candidate solutions for which fitness was estimated using the fitness model. Speed-up is also computed, which is equal to the factor by which the number of evaluations decreases compared to the case with no inheritance.

\begin{figure}[t]
\begin{center}
\subfigure[Number of evaluations.]{\epsfig{file=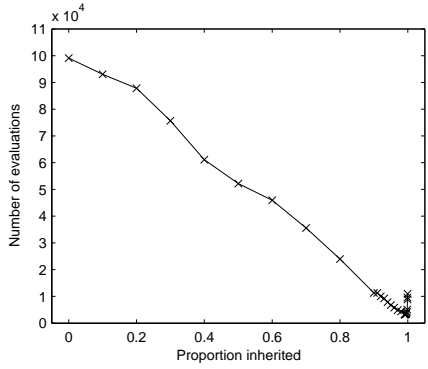,width=2.25in}}
\hspace{2ex}
\subfigure[Speed-up.]{\epsfig{file=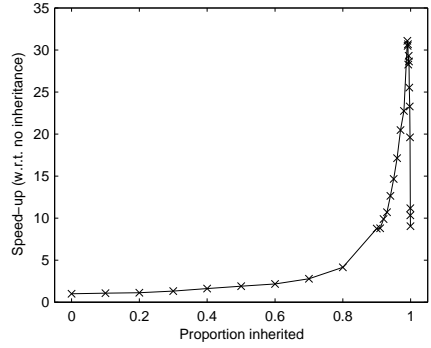,width=2.25in}}
\end{center}
\caption{Results on a 50-bit onemax.}
\label{figure-results-onemax}
\end{figure}

\begin{figure}[t]
\begin{center}
\subfigure[Number of evaluations.]{\epsfig{file=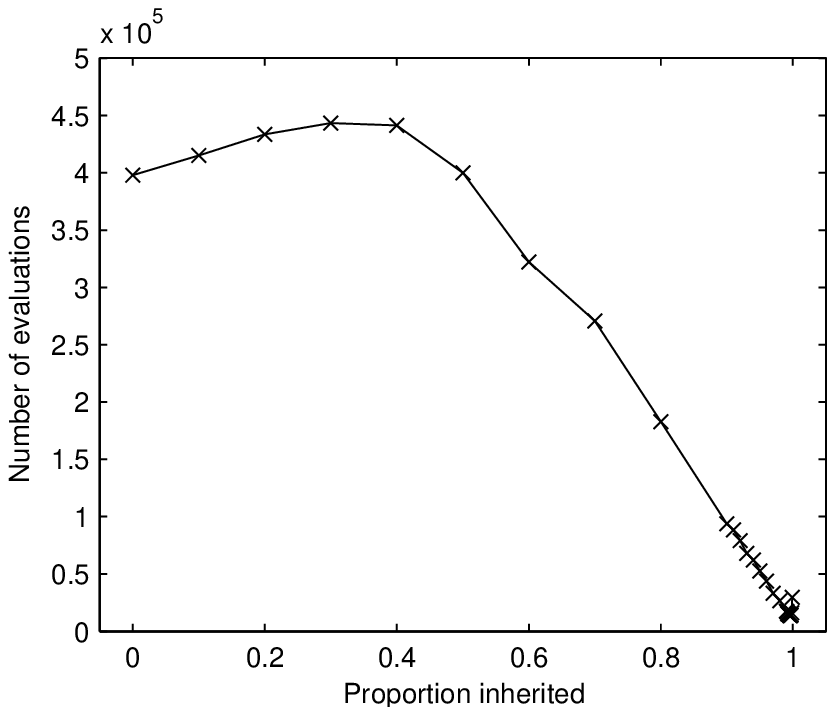,width=2.25in}}
\hspace{2ex}
\subfigure[Speed-up.]{\epsfig{file=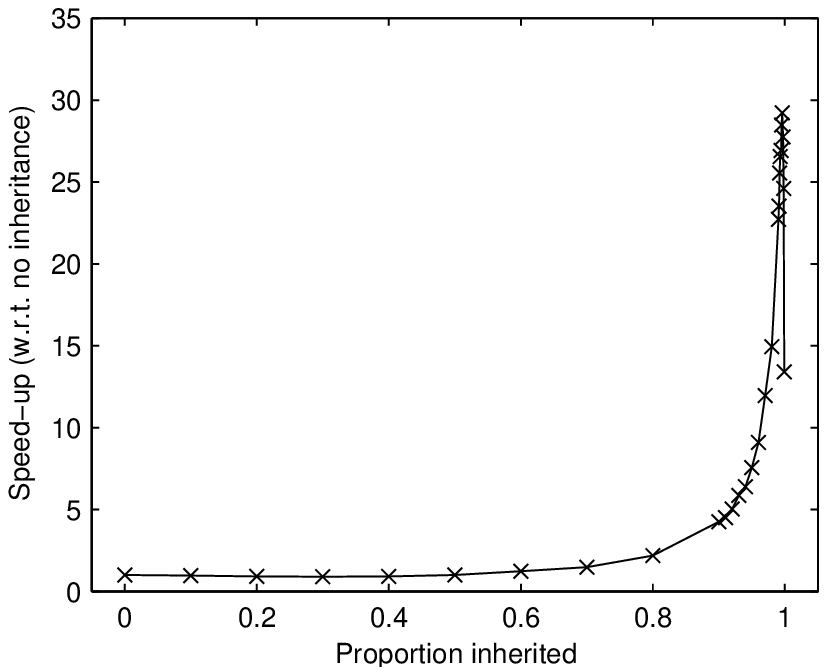,width=2.25in}}
\end{center}
\caption{Results on a concatenated trap consisting of 10 traps of order 4.}
\label{figure-results-trap4}
\end{figure}

\begin{figure}[t]
\begin{center}
\subfigure[Number of evaluations.]{\epsfig{file=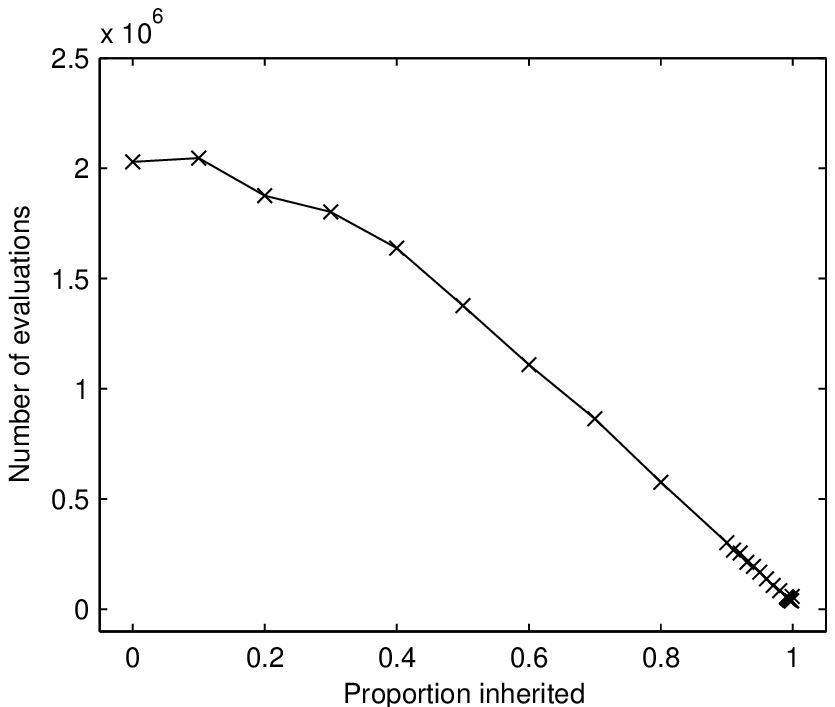,width=2.25in}}
\hspace{2ex}
\subfigure[Speed-up.]{\epsfig{file=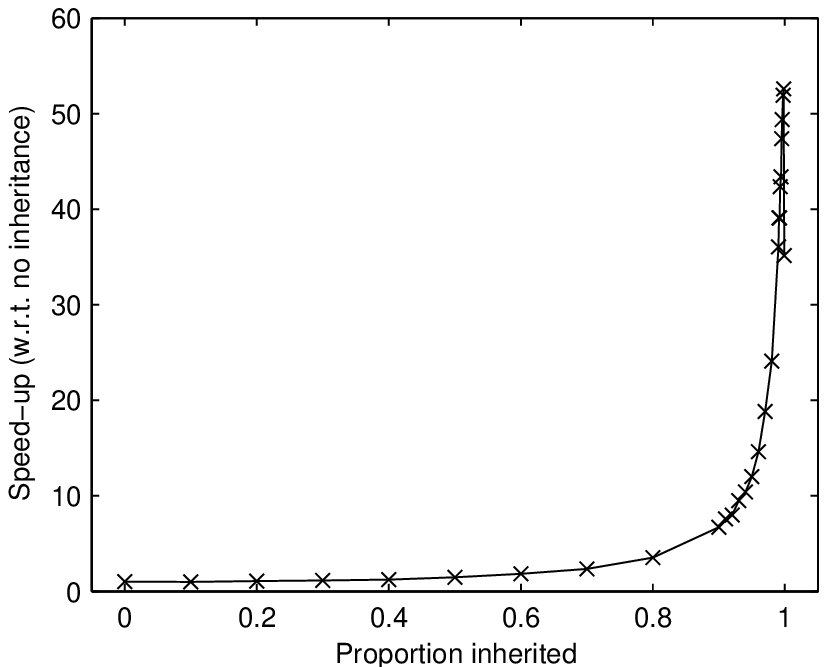,width=2.25in}}
\end{center}
\caption{Results on a concatenated trap consisting of 10 traps of order 5.}
\label{figure-results-trap5}
\end{figure}

The results on onemax, traps of order 4, and traps of order 5, are shown in figures~\ref{figure-results-onemax}, \ref{figure-results-trap4}, and~\ref{figure-results-trap5}. In all experiments, the number of actual fitness evaluations decreases with the inheritance proportion and it reaches the optimum when the proportion of candidate solutions for fitness inheritance is more than $99\%$. That means that considering only the actual fitness evaluations, evaluating less than $1\%$ of candidate solutions with the actual fitness seems to be beneficial. The number of evaluations of the actual fitness can be decreased by a factor of more than 31 for onemax, 32 for the trap of order 4, and 53 for the trap of order 5. Although the actual savings depend on the problem considered, it can be expected that fitness inheritance enables significant reduction of fitness evaluations on many problems because deceptive problems of bounded difficulty bound a large class of important problems.

Considering only the actual fitness evaluations ignores time complexity of selection, model construction, generation of new candidate solutions, and fitness estimation. Combining these factors with the complexity estimate for the actual fitness evaluation can be used to compute the optimal proportion of candidate solutions to evaluate using fitness inheritance. Nonetheless, the results presented in this paper clearly indicate that using fitness inheritance in BOA can reduce the number of solutions that must be evaluated using the actual fitness function by a factor of 30 or more. Consequently, if fitness evaluation is a bottleneck, there is a lot of space for improvement using fitness inheritance in BOA. 

\section{Summary and conclusions}
\label{section-conclusions}
\vspace*{-1ex}
Fitness inheritance enables genetic and evolutionary algorithms to evaluate only a certain proportion of candidate solutions using the actual fitness function, while the fitness of remaining solutions is computed using a model of the fitness landscape updated on the fly. Using fitness models that can be updated and used efficiently can significantly speed up solution to problems where fitness evaluation is computationally expensive. 

This paper showed that while fitness inheritance yields only moderate speed-ups of about $20\%$ in simple GAs and UMDA, in BOA the benefits of using fitness inheritance become more significant. Due to rather large population-sizing requirements for creating an adequate probabilistic model of promising solutions in BOA, the number of actual function evaluations decreases even if less than $1\%$ of candidate solutions are evaluated using the actual fitness function, while the fitness of the remaining solutions is estimated using only its model. That is an important result, because BOA and other advanced PMBGAs often require large populations, and evaluating large populations can become intractable for problems with computationally expensive fitness evaluation. 

Increasing the proportion of candidate solutions evaluated using a fitness model results in greater population-sizing requirements, and the optimal inheritance proportion depends on the complexity of building and sampling the model of promising solutions as well as that of evaluating solutions using the actual fitness function. The good news is that the more complex the evaluation function, the higher proportions of candidate solutions can be evaluated using the model of fitness instead of the actual fitness function. 

An important topic for future work is to incorporate fitness inheritance in presence of niching, which can lead to accumulation of candidate solutions whose fitness is overestimated. Resolving this problem would enable the use of fitness inheritance in the hierarchical BOA (hBOA)~\cite{Pelikan:01*,Pelikan:03b}, which combines BOA with local structures and niching. Another important topic is to develop theory that extends theoretical work on fitness inheritance in UMDA to BOA and other competent GAs. Finally, it is important to apply the proposed fitness inheritance model to solve challenging real-world problems with expensive fitness evaluation. 

\vspace*{-1ex}
\section*{Acknowledgments}
\vspace*{-1ex}
The authors would like to thank David E. Goldberg for discussions and comments. A part of this work was supported by the Research Award at the University of Missouri at St. Louis and the Research Board at the University of Missouri. Most experiments were done on Asgard cluster at the Institute of Theoretical Physics at the Swiss Federal Institute of Technology (ETH) Z\"{u}rich. The hBOA software, used by Pelikan, was developed by Martin Pelikan and David E. Goldberg at the University of Illinois at Urbana-Champaign. 

\vspace*{-2ex}
\bibliographystyle{apa-uiuc}
\bibliography{mybib}

\begin{thebibliography}{}

\bibitem[\protect\citeauthoryear{Ackley}{Ackley}{1987}]{Ackley:87b}
Ackley, D.~H. (1987).
\newblock An empirical study of bit vector function optimization.
\newblock {\em Genetic {A}lgorithms and {S}imulated {A}nnealing\/}, 170--204.

\bibitem[\protect\citeauthoryear{Buntine}{Buntine}{1991}]{Buntine:91}
Buntine, W.~L. (1991).
\newblock Theory refinement of {B}ayesian networks.
\newblock {\em Proceedings of the Uncertainty in Artificial Intelligence
  (UAI-91)\/}, 52--60.

\bibitem[\protect\citeauthoryear{Chickering, Heckerman, \& Meek}{Chickering
  et~al.}{1997}]{Chickering:97}
Chickering, D.~M., Heckerman, D., \& Meek, C. (1997).
\newblock {\em A {B}ayesian approach to learning {B}ayesian networks with local
  structure} (Technical Report MSR-TR-97-07).
\newblock Redmond, WA: Microsoft Research.

\bibitem[\protect\citeauthoryear{Cooper \& Herskovits}{Cooper \&
  Herskovits}{1992}]{Cooper:92}
Cooper, G.~F., \& Herskovits, E.~H. (1992).
\newblock A {B}ayesian method for the induction of probabilistic networks from
  data.
\newblock {\em Machine Learning\/},~{\em 9\/}, 309--347.

\bibitem[\protect\citeauthoryear{Deb \& Goldberg}{Deb \&
  Goldberg}{1994}]{Deb:94b}
Deb, K., \& Goldberg, D.~E. (1994).
\newblock Sufficient conditions for deceptive and easy binary functions.
\newblock {\em Annals of Mathematics and Artificial Intelligence\/},~{\em
  10\/}, 385--408.

\bibitem[\protect\citeauthoryear{Friedman \& Goldszmidt}{Friedman \&
  Goldszmidt}{1999}]{Friedman:99}
Friedman, N., \& Goldszmidt, M. (1999).
\newblock Learning {B}ayesian networks with local structure.
\newblock In Jordan, M.~I. (Ed.), {\em Graphical models} (pp.\  421--459).
  Cambridge, MA: MIT Press.

\bibitem[\protect\citeauthoryear{Goldberg}{Goldberg}{1989}]{Goldberg:89d}
Goldberg, D.~E. (1989).
\newblock {\em Genetic algorithms in search, optimization, and machine
  learning}.
\newblock Reading, MA: Addison-Wesley.

\bibitem[\protect\citeauthoryear{Heckerman, Geiger, \& Chickering}{Heckerman
  et~al.}{1994}]{Heckerman+al:94}
Heckerman, D., Geiger, D., \& Chickering, D.~M. (1994).
\newblock {\em Learning {B}ayesian networks: {T}he combination of knowledge and
  statistical data} (Technical Report MSR-TR-94-09).
\newblock Redmond, WA: Microsoft Research.

\bibitem[\protect\citeauthoryear{Holland}{Holland}{1975}]{Holland:75a}
Holland, J.~H. (1975).
\newblock {\em Adaptation in natural and artificial systems}.
\newblock Ann Arbor, MI: University of Michigan Press.

\bibitem[\protect\citeauthoryear{Howard \& Matheson}{Howard \&
  Matheson}{1981}]{Howard:81}
Howard, R.~A., \& Matheson, J.~E. (1981).
\newblock Influence diagrams.
\newblock In Howard, R.~A., \& Matheson, J.~E. (Eds.), {\em Readings on the
  principles and applications of decision analysis}, Volume~II (pp.\
  721--762). Menlo Park, CA: Strategic Decisions Group.

\bibitem[\protect\citeauthoryear{Pearl}{Pearl}{1988}]{Pearl:88}
Pearl, J. (1988).
\newblock {\em Probabilistic reasoning in intelligent systems: {N}etworks of
  plausible inference}.
\newblock San Mateo, CA: Morgan Kaufmann.

\bibitem[\protect\citeauthoryear{Pelikan}{Pelikan}{2002}]{Pelikan:thesis}
Pelikan, M. (2002).
\newblock {\em Bayesian optimization algorithm: {F}rom single level to
  hierarchy}.
\newblock Doctoral dissertation, University of Illinois at Urbana-Champaign,
  Urbana, IL.
\newblock Also {IlliGAL Report No.} 2002023.

\bibitem[\protect\citeauthoryear{Pelikan \& Goldberg}{Pelikan \&
  Goldberg}{2001}]{Pelikan:01*}
Pelikan, M., \& Goldberg, D.~E. (2001).
\newblock Escaping hierarchical traps with competent genetic algorithms.
\newblock {\em Proceedings of the {G}enetic and {E}volutionary {C}omputation
  {C}onference ({GECCO}-2001)\/}, 511--518.
\newblock Also {IlliGAL Report No.} 2000020.

\bibitem[\protect\citeauthoryear{Pelikan \& Goldberg}{Pelikan \&
  Goldberg}{2003}]{Pelikan:03b}
Pelikan, M., \& Goldberg, D.~E. (2003).
\newblock A hierarchy machine: {L}earning to optimize from nature and humans.
\newblock {\em Complexity\/},~{\em 8\/}(5).

\bibitem[\protect\citeauthoryear{Pelikan, Goldberg, \& Cant{\'{u}}-Paz}{Pelikan
  et~al.}{1999}]{Pelikan:99a}
Pelikan, M., Goldberg, D.~E., \& Cant{\'{u}}-Paz, E. (1999).
\newblock {BOA}: The {B}ayesian optimization algorithm.
\newblock {\em Proceedings of the {G}enetic and {E}volutionary {C}omputation
  {C}onference ({GECCO}-99)\/},~{\em I\/}, 525--532.
\newblock Also {IlliGAL Report No.} 99003.

\bibitem[\protect\citeauthoryear{Pelikan, Goldberg, \& Lobo}{Pelikan
  et~al.}{2002}]{Pelikan:02}
Pelikan, M., Goldberg, D.~E., \& Lobo, F. (2002).
\newblock A survey of optimization by building and using probabilistic models.
\newblock {\em Computational Optimization and Applications\/},~{\em 21\/}(1),
  5--20.
\newblock Also {IlliGAL Report No.} 99018.

\bibitem[\protect\citeauthoryear{Pelikan, Goldberg, \& Sastry}{Pelikan
  et~al.}{2001}]{Pelikan:01a*}
Pelikan, M., Goldberg, D.~E., \& Sastry, K. (2001).
\newblock Bayesian optimization algorithm, decision graphs, and {O}ccam's
  razor.
\newblock {\em Proceedings of the {G}enetic and {E}volutionary {C}omputation
  {C}onference ({GECCO}-2001)\/}, 519--526.
\newblock Also {IlliGAL Report No.} 2000020.

\bibitem[\protect\citeauthoryear{Sastry, Goldberg, \& Pelikan}{Sastry
  et~al.}{2001}]{Sastry:01}
Sastry, K., Goldberg, D.~E., \& Pelikan, M. (2001).
\newblock Don't evaluate, inherit.
\newblock {\em Proceedings of the {G}enetic and {E}volutionary {C}omputation
  {C}onference ({GECCO}-2001)\/}, 551--558.
\newblock Also {IlliGAL Report No.} 2001013.

\bibitem[\protect\citeauthoryear{Schwarz}{Schwarz}{1978}]{Schwarz78a}
Schwarz, G. (1978).
\newblock Estimating the dimension of a model.
\newblock {\em The Annals of Statistics\/},~{\em 6\/}, 461--464.

\bibitem[\protect\citeauthoryear{Smith, Dike, \& Stegmann}{Smith
  et~al.}{1995}]{SmithR:94}
Smith, R.~E., Dike, B.~A., \& Stegmann, S.~A. (1995).
\newblock Fitness inheritance in genetic algorithms.
\newblock {\em Proceedings of the ACM Symposium on Applied Computing\/},
  345--350.

\bibitem[\protect\citeauthoryear{Zheng, Julstrom, \& Cheng}{Zheng
  et~al.}{1997}]{Zheng:97}
Zheng, X., Julstrom, B., \& Cheng, W. (1997).
\newblock Design of vector quantization codebooks using a genetic algorithm.
\newblock In {\em Proceedings of the International Conference on Evolutionary
  Computation ({ICEC}-97)} (pp.\  525--529). Picataway, NJ: IEEE Press.

\end{thebibliography}

\end{document}